# ADABOOST WITH "KEYPOINT PRESENCE FEATURES" FOR REAL-TIME VEHICLE VISUAL DETECTION

**Taoufik Bdiri, Fabien Moutarde, Nicolas Bourdis and Bruno Steux**

Robotics Laboratory (CAOR), Mines ParisTech
60 Bd Saint-Michel, F-75006 PARIS, FRANCE
Tel.: (33) 1-40.51.92.92, Fax: (33) 1.43.26.10.51
`Fabien.Moutarde@mines-paristech.fr`

**ABSTRACT**

We present promising results for real-time vehicle visual detection, obtained with adaBoost using new original "keypoints presence features". These weak-classifiers produce a boolean response based on presence or absence in the tested image of a "keypoint" (~ a SURF interest point) with a descriptor sufficiently similar (i.e. within a given distance) to a reference descriptor characterizing the feature. A first experiment was conducted on a public image dataset containing lateral-viewed cars, yielding 95% recall with 95% precision on test set. Moreover, analysis of the positions of adaBoost-selected keypoints show that they correspond to a specific part of the object category (such as "wheel" or "side skirt") and thus have a "semantic" meaning.

## INTRODUCTION AND RELATED WORKS

Efficient and reliable detection of surrounding moving objects (such as pedestrians and vehicles) is one of the key features for enhancing safety in intelligent vehicles. It is particularly interesting to be able to properly detect laterally incoming cars that could lead to lateral collisions. A module implementing reliable (particularly with very low false alarm rate) visual detection of laterally incoming cars could for instance be used for developing useful Advanced Driving Assistance Systems (ADAS) such as a Lateral Collision Warning (LCW) system.

We present here a new method for the visual recognition and detection of a given object type, with a first test application for laterally-viewed cars. Many techniques have been proposed for visual object detection and classification (see e.g. [3] for a review of some of the state-of-the-art methods for pedestrian detection, which is the most challenging). Of the various machine-learning approaches applied to this problem, only few are able to process videos in real-time. Among those, the boosting algorithm with feature selection was successfully extended to machine-vision by Viola & Jones [2]. The adaBoost algorithm was introduced in 1995 by Y. Freund and R. Shapire [1], and its principle is to build a *strong classifier*, assembling weighted weak classifiers, those being obtained iteratively by using successive weighting of the examples in the training set. Most published works using adaBoost for visual object class detection are using the Haar-like features initially proposed by Viola & Jones for face and pedestrian detection.

However, adaBoost outcome may strongly depend on the family of features from which the weak classifiers are drawn. Recently, several teams [4][5] have reported interesting results with boosting using other kinds of features directly inspired from the Histogram of Oriented Gradient (HOG) approach. Our lab has been successfully investigating boosting with pixel-comparison-based features named "control-points" (see [6] for original proposal, and [7] for recent results with a new variant).

To our knowledge, the idea of using interest point descriptors as boosting features was first proposed by Opelt et al. in [8], but it was in a more general framework, and they were considering SIFT points and descriptors [9] which are quite slow to compute, compared to the SURF points and descriptors [10]. In the present work we investigate boosting of "keypoint presence features", where "keypoint" are a variant of SURF points implemented in our lab, and already successfully applied to real-time person re-identification [11].

## CAMELLIA "KEYPOINTS"

The interest point detection and descriptor computation is performed using "key-points" functions available in the Camellia image processing library (http://camellia.sourceforge.net). These Camellia key-points detection and descriptor functions – named CamKeypoints - implement a variant of SURF [10]. SURF itself is an extremely efficient method (thanks to the use of integral images) inspired from the more classic and widely used interest point detector and descriptor SIFT [9].

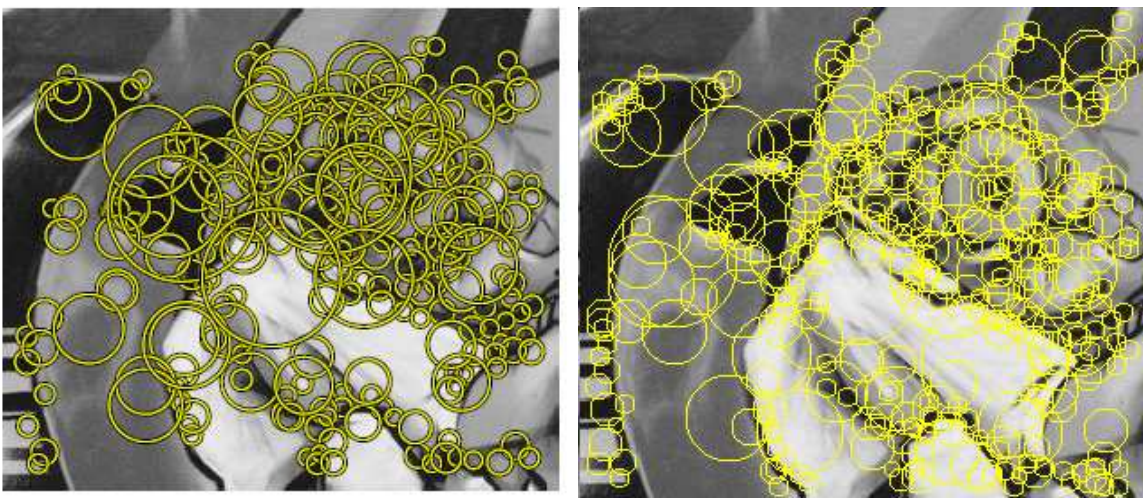

*Figure 1 - SURF interest points (left) v.s. Camellia keypoints (right); they are very similar except for voluntary suppression of multiple imbricated blobs at different scales (cf. upper left).*

As for SURF interest points, the detection of Camellia keypoints is a "blob detector" based on finding local Hessian maxima, those being efficiently obtained by approximating second order derivatives with box filters computed with integral image. Our keypoints are however not exactly the same as SURF points (as can be seen on figure 1), in particular because multiple imbricated blobs at various scales are voluntarily avoided. In contrary to SURF and SIFT, CamKeypoint scale selection is not based on overlapping octaves, but on a set of discrete scales from which the scale of a keypoint is derived by quadratic interpolation. This speeds up the keypoints detection compared to SURF by a factor of 2, without sacrificing the quality of scale information, as was shown by some experiments.

The descriptor used for each Camellia keypoint is similar to the SURF descriptor : an image patch corresponding to the keypoint location and scale is divided in 4x4=16 sub-regions, on each of which are efficiently computed (by using integral image approach) the following 4 quantities:

$$\sum dx$$
$$\sum |dx|$$
$$\sum dy$$
$$\sum |dy|$$

where *dx* and *dy* are respectively the horizontal and vertical gradient. The total descriptor size is therefore 16 x 4 = 64. In order to avoid all boundary effects in which the descriptor abruptly changes when a keypoint physically changes, bi-linear extrapolation is used to distribute each of the 4 quantities above into 4 sub-regions. Experiments have shown that this really improves the quality of the descriptor wrt. SURF. In addition to this, CamKeypoints support color images by adding 32 elements of gradient information by color channel (U and V) to the signature, resulting in a 128 descriptor size for color descriptors.

Another main difference between Camellia Keypoints and SURF lies in that the Camellia implementation uses integer-only computations – even for the scale interpolation –, which makes it even faster than SURF, and particularly well-suited for potential embedding in camera hardware. SIFT and SURF make extensive use of floating point computations, which makes these algorithms power hungry.

**BOOSTING "KEYPOINTS PRESENCE " FOR CATEGORIZATION**

The rationale of boosting "keypoint presence features" for image categorization is that it should be possible, for a given object category, to determine a set of characteristic interest points whose *simultaneous* presence would be representative of that particular category. This is similar in spirit, but with a completely different algorithm, to the "part based" approach proposed by [12].

Our object recognition approach uses the same general feature-selecting boosting framework as pioneered by Viola&Jones in [2]. The originality of our work is to define and use as weak classifiers a new original feature family, instead of Haar features. This new feature type is a weak classifier that answers positively on an image if and only if, among all the Camellia keypoints detected in the image, there is at least one of them whose descriptor is similar enough to the "reference keypoint descriptor" associated with the weak-classifier.

More formally, each "keypoint presence" weak-classifier is defined by a keypoint SURF descriptor D (in $\Re^{64}$) and a descriptor difference threshold scalar value d. This weak-classifier h(D,d,I) answers positively on an image I if and only if I contains at least one keypoint whose descriptor D' is such that |D-D'|<d, where the "sum of absolute difference" (SAD) L1-distance is used: if the descriptors of two keypoints $K_1$ and $K_2$ are respectively {$D_1$[i], i = 1…64} and {$D_2$[i], i = 1…64}, then, the distance between K1 and K2 is given by equation 1 below:

$$\text{Dist}(K_1, K_2) = \sum_{i=1}^{64} \text{abs}(D_1[i] - D_2[i]) \quad \text{(Eq. 1)}$$

The training method is the standard feature-selecting adaBoost algorithm, in which the descriptor D is chosen among all descriptors found in *positive* example images. The threshold value d is chosen by considering the matrix M of distances Mij between positive keypoint Ki and all images Ij, this distance being defined as the smallest descriptor difference |D(Ki)-D(Kpj)| for all keypoints Kpj found in image j.

More formally, let the training set be composed of positive images Ip1, Ip2, …., and of negative images In1, In2, …We first apply the Camellia keypoint detector on all *positive* images Ip1, Ip2, …, and build the "positive keypoints set" Spk = {$K_1$, $K_2$, $K_3$, …, $K_Q$} as the union of all Camellia keypoints detected on any positive examples of the training set. The adaBoost feature-selection has to select, at each boosting step, a particular "keypoint presence" weak classifier defined by a 64D descriptor and a scalar threshold. The descriptor will be chosen among those of positive keypoints collected in Spk.

In order to choose a threshold value, we apply keypoints detection on all negative images as well, so that we can compare descriptors of the positive keypoints in Spk to descriptors of all keypoints found in training images. We define the "distance" between any given keypoint K and any given image I as the smallest descriptor difference between K and all keypoints KIj found in image I:

$$\text{dist}(K,I) = \min_{\text{KIj keypoint found in image I}} \{ \text{dist}(K, KI_j) \} \quad \text{(Eq. 2)}$$

where dist(K,KIj) is the SAD of descriptors as defined in equation (1). This allows us to build a matrix M of distances between positive keypoints and all training images, where Mij = dist(Ki , Ij). As illustrated on figure 2, this QxN matrix (with Q the number of positive keypoints and N the number of training images) has at least one zero on each line, on the column corresponding to the positive image in which the keypoints was found.

We execute a growing sorting of the distance matrix M, row by row, and then we take the *middle of each two successive distances* in the sorted matrix to build the set {$T_{ik}$, k=1,…,N} of candidate threshold values for a feature testing presence of the corresponding positive keypoint $K_i$.

$$\begin{pmatrix} & I_{p1} & I_{p2} & I_{p3} & I_{n1} & I_{n2} & I_{n3} \\ 0 & x & x & x & x & x \\ 0 & x & x & x & x & x \\ 0 & x & x & x & x & x \\ x & 0 & x & x & x & x \\ x & 0 & x & x & x & x \\ x & 0 & x & x & x & x \\ x & x & 0 & x & x & x \\ x & x & 0 & x & x & x \\ x & x & 0 & x & x & x \\ x & x & 0 & x & x & x \end{pmatrix}$$

*Figure 2 - Matrix of distances between keypoints found on positive image examples (one for each row) and all N training images (positives and negatives, one image for each column)*

At each boosting step, we choose among all $(K_i, T_{ik})$ couples the one that gives the lowest weighted error on the training set: $(i^*, k^*) = \text{argmin}_{ik} \left( \sum_{j=1}^{N} w_j |h(K_i, T_{ik}, I_j) - l_j| \right)$, and the selected weak classifier is $h(K_{i^*}, T_{i^*k^*}, .)$.

## EXPERIMENT ON LATERAL-VIEWED CAR DATASET

For a first evaluation of our approach, we used the publicly available (http://l2r.cs.uiuc.edu/~cogcomp/Data/Car/) lateral-car dataset collected by Agarwal et al. [12]. This database contains 550 positive images and 500 negative images. For training, we use 352 positive images, and 322 negative images, the rest being used as a test set for evaluation. Note that the partition between training and testing subset is random. Some examples from the training set are shown on figure 3.

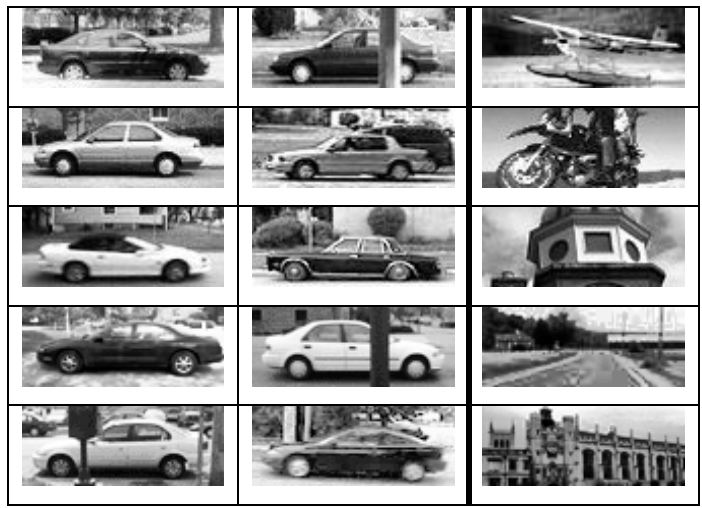

*Figure 3 - Some positive (2 left columns) and negative (right column) examples from the training set*

Figure 4 shows the typical error evolution during adaBoost training: as is usual with boosting, the training error quickly falls to zero, and the error on test set continues to diminish afterwards. This shows that boosting by assembling features extracted from our new "keypoint presence" family does work and allow to build a strong classifier able to discriminate a given object category. On this particular case, there seems to be no clear improvement on test dataset for boosting steps T>150.

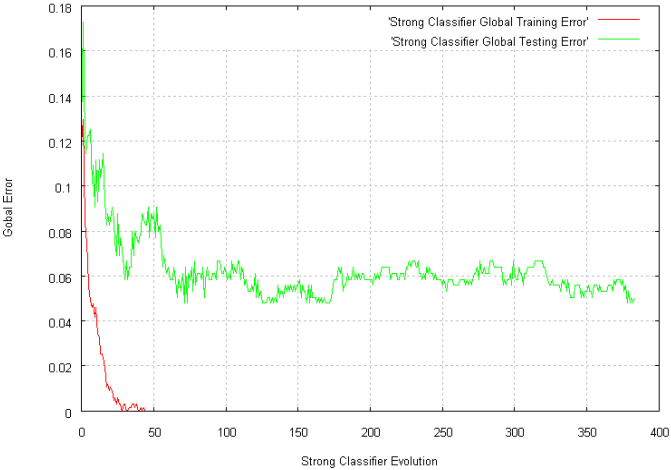

*Figure 4 - Typical evolution, during successive boosting steps,
of errors on training and test*

Figure 5 shows the precision-recall curve, computed on the independent test set, for boosted strong classifiers with respectively 10 and 300 "keypoint presence" weak-classifiers assembled. The classification result is very good, with a recall (percentage of lateral cars recognized) of ~95%, for a precision (percentage of true lateral cars among all test images declared positive by the classifier) of ~95%.

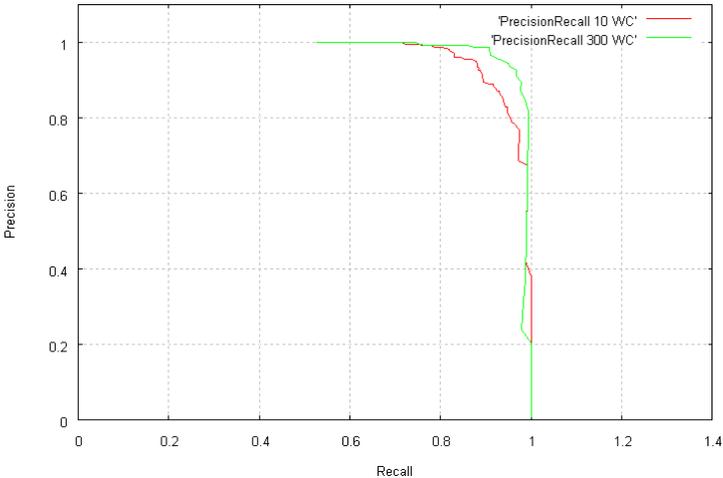

*Figure 5 - Precision-recall curve computed on test set,
for strong boosted classifier assembling 10 and 300 weak classifiers.*

There are several motivations for our new feature type. One is that a classifier based on the simultaneous presence of several characteristic keypoints matches the intuition we can have on how human do categorize image by spotting some characteristic parts. In order to check if our adaBoost-selected keypoints make sense from this point of view, we decided to check on positive images where are located the "positively responding keypoints" for a given feature of the strong classifier. Figure 6 illustrates the positions of all keypoints, cumulated on all positive example images, that are within the descriptor distance threshold of one given adaBoost-selected keypoints. This clearly shows that the keypoints selected correspond to specific parts of the object category, such as the wheels or the side skirt, which means they have a semantic signification relative to the object category.

| Weak Classifier | Image 1 | Specificity |
|---|---|---|
| 40 | 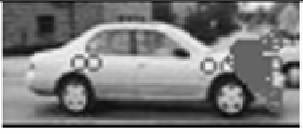 | Keypoints concentrated on the rectangular side of the car |
| 230 | 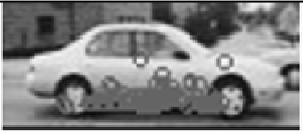 | Keypoints concentrated under the car between the two wheels |
| 300 | 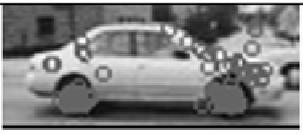 | Keypoints concentrated on wheels |

*Figure 6 - Position of adaBoost positively responding keypoints, cumulated on all positive example images: each selected keypoint seem to correspond to a specific part of the car.*

Another motivation for these new kind of adaBoost features is that, by nature of the features, it is possible to derive the localizations in the image of objects of the searched category quite straightforwardly by some kind of Hough-like method applied to the positions of object-category-specific keypoints, thus making costly window-scanning unnecessary.

A preliminary result of keypoints filtering followed by object detection and localization, applied on a real on-board video, is illustrated on figure 7. The keypoint filtering uses a specific keypoint classifier trained to discriminate between "lateral car keypoints" and "background keypoints". From these remaining keypoints, those that are compatible with one of our adaBoost-selected weak-classifiers are used to derive candidate bounding-boxes by applying a Hough-like method

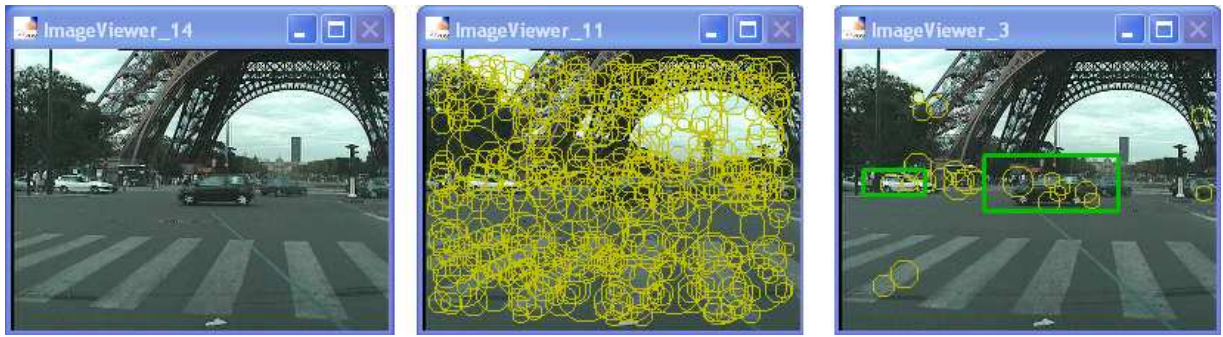

*Figure 7 - First detection test on a video: the processed frame on left side, all keypoints on middle, and on the right side only keypoints classified as "lateral-car keypoints" and the bounding boxes obtained with our Hough-like localization method.*

## CONCLUSIONS, DISCUSSION AND PERPECTIVES

We have presented a first successful test of boosting "keypoint presence features", applied to lateral car recognition, yielding 95% recall with 95% precision on test set. Moreover, analysis of the positions of adaBoost-selected keypoints show that they correspond to a specific part of the object category (such as "wheel" or "side skirt") and thus have a "semantic" meaning.

Regarding the performance attained, it is important to note that in the potential application (a Lateral Collision Warning system), it is not too problematic to miss some of the cars, but what is most important is to ensure a low false alarm rate. Therefore a 95% recognition rate is quite sufficient. The 95% precision could seem to imply too many false alarms, but one should keep in mind that we report here only rate computed on a still images database, in which the negative examples can be more confusing than background on an empty road or street. Also, the target application would analyze a video, so that a temporal filtering would most probably get rid of most false alarms, which would probably not arise on all successive frames, contrary to true laterally incoming cars.

Perspectives include tests on various other datasets, including for other object categories such as pedestrians (for which a preliminary test was encouraging). If tests on other dataset for other categories are also successful, then this would imply that our method is quite general, and could be used for recognition of various types of potential obstacle.

Another important work underway is the improvement and optimization of our object-localization method based on the analysis of positions of positively-responding keypoints, that allows the detection step to be done without any tedious window-scanning, contrary to most existing detection object detection algorithms. Finally, we think the recognition performance of our method could be further improved by exploiting the relative positions of keypoints, instead of only their presence.